\pdfoutput=1

\documentclass[11pt]{article}

\usepackage[]{acl} 

\usepackage{times}
\usepackage{latexsym}

\usepackage[T1]{fontenc}

\usepackage[utf8]{inputenc}

\usepackage{microtype}
\usepackage{multirow}
\usepackage{tikz}
\usepackage{tipa}
\usepackage{amssymb}
\usepackage{pifont}
\newcommand{\cmark}{\ding{51}}%
\newcommand{\xmark}{\ding{55}}%
\usepackage{xspace}
\newcommand*{\yoruba}{Yor\`ub\'a\xspace}
\usepackage[T1]{fontenc}
\usepackage{amsmath,graphicx}
\usepackage{amsfonts}

\title{UCAS-IIE-NLP at SemEval-2023 Task 12: Enhancing Generalization of Multilingual BERT for Low-resource Sentiment Analysis}

\author{Dou Hu$^{1,2}$ 
        \and Lingwei Wei$^{1,2}$
        \and Yaxin Liu$^{1,2}$
        \and Wei Zhou$^{1}$
        \and Songlin Hu$^{1,2}$
         \\
        $^{1}$ Institute of Information Engineering, Chinese Academy of Sciences \\
        $^{2}$ School of Cyber Security, University of Chinese Academy of Sciences  \\
        \texttt{\{hudou, weilingwei, liuyaxin, zhouwei, husonglin\}@iie.ac.cn} \\
}

\begin{document}
\maketitle
\begin{abstract}

This paper describes our system designed for SemEval-2023 Task 12: Sentiment analysis for African languages. The challenge faced by this task is the scarcity of labeled data and linguistic resources in low-resource settings. To alleviate these, we propose a generalized multilingual system \textbf{SACL-XLMR} for sentiment analysis on low-resource languages. Specifically, we design a lexicon-based multilingual BERT to facilitate language adaptation and sentiment-aware representation learning. 
Besides, we apply a supervised adversarial contrastive learning technique to learn sentiment-spread structured representations and enhance model generalization. Our system achieved competitive results, largely outperforming baselines on both multilingual and zero-shot sentiment classification subtasks. Notably, the system obtained the \textbf{1st} rank on the zero-shot classification subtask in the official ranking. Extensive experiments demonstrate the effectiveness of our system.

\end{abstract}

\section{Introduction}

Sentiment analysis is a critical aspect of natural language processing with numerous applications, including public opinion monitoring \citep{info:doi/10.2196/21978}, healthcare services \citep{info:doi/10.2196/16023}, and recommendation systems \citep{DBLP:conf/ksem/HuWZHFH21}. However, performing sentiment analysis in low-resource languages poses significant challenges, including the scarcity of labeled data and linguistic resources, as well as the diversity of languages and dialects \citep{DBLP:journals/air/LoCCC17,DBLP:journals/fgcs/OueslatiCHO20}. In SemEval-2023 Task 12 \citep{muhammadSemEval2023}, the focus is on sentiment analysis for African languages in Twitter, which further exacerbates the challenges due to the presence of tone, code-switching, and digraphia phenomena \cite{adebara-abdul-mageed-2022-towards}.

Although multilingual pre-trained language models (multilingual PTMs) \citep{DBLP:conf/nips/ConneauL19,DBLP:conf/acl/ConneauKGCWGGOZ20} have shown potential in cross-lingual transfer learning compared to monolingual PTMs \citep{DBLP:conf/naacl/DevlinCLT19,DBLP:conf/emnlp/0001HDZJMS22}, they have limitations in capturing nuances and cultural differences within a language, especially in the context of dialects and regional variations.

In this paper, we propose a generalized multilingual system named \textbf{SACL-XLMR} to address these limitations and enhance the generalization of multilingual PTMs for under-represented languages, particularly African languages. Our system leverages a lexicon-based multilingual BERT model to facilitate language adaptation and sentiment-aware representation learning. Additionally, we apply a supervised adversarial contrastive learning (SACL) technique \citep{hu2023supervised} to learn sentiment-spread structured representations and enhance model generalization.

We present the details of the proposed system and evaluate its performance on SemEval-2023 Task 12. Our system achieves remarkable performance, outperforming baselines by \textbf{+1.1\%} weighted-F1 score on multilingual sentiment classification subtask and by \textbf{+2.8\%} weighted-F1 score on zero-shot sentiment classification subtask in the AfriSenti-SemEval datasets \citep{muhammad2023afrisenti}. 
Moreover, following the AfriSenti SemEval Prizes\footnote{\url{https://afrisenti-semeval.github.io/prizes/}} and the task description \citep{muhammadSemEval2023}, our system obtains the \textbf{1st} rank on the zero-shot classification subtask in the official ranking.
We conducted experiments to demonstrate the effectiveness of our approach, highlighting the potential of our system in overcoming the challenges of low-resource sentiment analysis.

\section{Background}

\subsection{Task and Data Description}

\begin{table*}[t]
\centering
\resizebox{\linewidth}{!}{
\begin{tabular}{l|l|rrrr|l|c|c}
\hline
\multicolumn{1}{c|}{ISO Code} 
& \multicolumn{1}{c|}{Language} 
& \multicolumn{1}{c}{Total} & \multicolumn{1}{c}{Train} & \multicolumn{1}{c}{Val} & \multicolumn{1}{c|}{Test} 
& \multicolumn{1}{c|}{Subregion} 
& \multicolumn{1}{c|}{Script} 
& \multicolumn{1}{c}{Lexicon} \\ \hline 
\texttt{amh} & Amharic & 9,483 & 5,985 & 1,498 & 2,000 
& East Africa & Ethiopic & \xmark \\ 
\texttt{arq} & Algerian Arabic/Darja & 3,062 & 1,652 & 415 & 959 
& North Africa & Arabic & \xmark \\ 
\texttt{hau} & Hausa & 22,155 & 14,173 & 2,678 & 5,304 
& West Africa & Latin & \cmark  \\ 
\texttt{ibo} & Igbo & 15,718 & 10,193 & 1,842 & 3,683 
& West Africa & Latin & \cmark  \\ 
\texttt{kin} & Kinyarwanda & 5,158 & 3,303 & 828 & 1,027 
& East Africa & Latin & \cmark \\ 
\texttt{ary} & Moroccan Arabic/Darija & 9,762 & 5,584 & 1,216 & 2,962
& Northern Africa & Arabic/Latin & \cmark \\ 
\texttt{pt-MZ} & Mozambican Portuguese & 7,495 & 3,064 & 768 & 3,663 
& Southeastern Africa & Latin & \xmark \\ 
\texttt{pcm} & Nigerian Pidgin & 10,559 & 5,122 & 1,282 & 4,155 
& West Africa & Latin & \xmark \\ 
\texttt{orm} & Oromo & 2,494 & - & 397 & 2,097
& East Africa & Latin & \cmark \\ 
\texttt{swa} & Swahili & 3,014 & 1,811 & 454 & 749
& East Africa & Latin & \xmark \\ 
\texttt{tir} & Tigrinya & 2,400 & - & 399 & 2,001 
& East Africa & Ethiopic & \cmark \\ 
\texttt{twi} & Twi & 4,821 & 3,482 & 389 & 950 
& West Africa & Latin & \cmark \\ 
\texttt{tso} & Xitsonga & 1,264 & 805 & 204 & 255
& Southern Africa & Latin & \xmark \\ 
\texttt{yor} & \yoruba & 15,130 & 8,523 & 2,091 & 4,516 
& West Africa & Latin & \cmark \\ 
\hline
\end{tabular}
}
\caption{\label{dataset-statistic}
The statistics of the AfriSenti datasets. 
The train/validation sets of Oromo (\texttt{orm}) and Tigrinya (\texttt{tir}) are not used due to 
the zero-shot transfer setting used for evaluation.
Lexicon refers to a valid lexicon, which provides words or phrases that correspond to the predefined sentiment polarity.}
\end{table*}
The SemEval-2023 Task 12: Sentiment analysis for African languages 
(AfriSenti-SemEval) \citep{muhammadSemEval2023} is the first Afro-centric SemEval shared task for sentiment analysis in Twitter. It consists of three subtasks, i.e., monolingual, multilingual, and zero-shot sentiment classification. 
Brief descriptions of the last two subtasks that our team focuses on are as follows:
\begin{itemize}
 \item \textbf{Multilingual Sentiment Classification}. Given combined training data of multiple African languages, determine the polarity of a tweet on the combined test data of the same languages (positive, negative, or neutral). This subtask has only one track with 12 languages (Amharic, Algerian Arabic/Darja, Hausa, Igbo, Kinyarwanda, Moroccan Arabic/Darija, Mozambican Portuguese, Nigerian Pidgin, Swahili, Twi, Xitsonga, and \yoruba),
 i.e., a multilingual track with 12 African languages.
 \item \textbf{Zero-Shot Sentiment Classification}. Given unlabelled tweets in two African languages (Tigrinya and Oromo), leverage any or all available training datasets of source languages (12 African languages in the multilingual track) to determine the sentiment of a tweet in the two target languages. This task has two tracks, i.e., a zero-shot Tigrinya track and a zero-shot Oromo track.
\end{itemize}

The AfriSenti datasets\footnote{\url{https://github.com/afrisenti-semeval}} \citep{muhammad2023afrisenti} are a collection of multilingual Twitter datasets that consist of 110,000+ tweets in 14 low-resource African languages from four language families for sentiment analysis.
The statistics of each monolingual tweet datasets are reported in Table~\ref{dataset-statistic}.
The datasets involve tweets labeled with three sentiment classes (positive, negative, neutral).
Each tweet is annotated by three native speakers following the sentiment annotation guidelines \citet{mohammad-2016-practical} and 
the final label for each tweet is determined by majority voting \citep{DBLP:journals/tacl/DavaniDP22}. If a tweet conveys both a positive and negative sentiment, the stronger sentiment should be chosen.

\section{Related Work}
\subsection{Sentiment Analysis}
Sentiment analysis has evolved from lexicon-based approaches to more advanced machine learning and deep learning-based methods \citep{MEDHAT20141093}. 
Previous works in sentiment analysis have focused on various levels of granularity, such as aspect \citep{DBLP:conf/semeval/PontikiGPPAM14}, sentence \citep{DBLP:conf/acl/HuWH20}, and document \citep{DBLP:conf/pkdd/WeiHZTZWHH20}, as well as different modalities \citep{DBLP:conf/emnlp/ZadehCPCM17,DBLP:conf/icassp/HuHWJM22} and languages \citep{DBLP:journals/ir/BoiyM09,DBLP:journals/csl/BalahurT14}.

\subsection{Low-resource Sentiment Analysis}
Despite the success of polarity classification in high-resource languages, noisy user-generated data in under-represented languages presents a challenge \citep{DBLP:conf/coling/YimamAAB20}. 
Recently, several studies have proposed approaches for sentiment analysis on low-resource languages \citep{DBLP:journals/air/LoCCC17,DBLP:conf/coling/YimamAAB20}.
Besides, \citet{moudjari-etal-2020-algerian,adebara-abdul-mageed-2022-towards,muhammad2023afrisenti} have relied on manual annotation by native speakers or expert annotators to build sentiment analysis datasets in low-resource languages.

\begin{figure*}[t]
    \centering
    \includegraphics[width=0.618\linewidth]{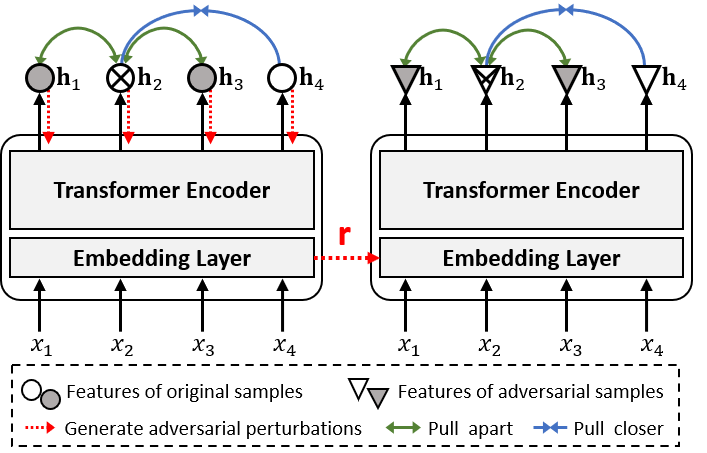} 
    \caption{Overall architecture of our SACL-XLMR.
    Given a batch of training samples, a multilingual BERT is used to learn contextual representations of the input sentences. We take the $\boldsymbol{\times}$-marked utterance as an example to show the objective of SACL.
    $r$ means adversarial perturbations that put on the embedding layer of BERT.
    }
    \label{fig:model}
\end{figure*}

\section{System Overview}

In this section, we describe our system adopted in SemEval-2023 Task 12, where we design a generalized multilingual system named \textbf{SACL-XLMR} for sentiment analysis on low-resource languages.

\subsection{Model Architecture}
The network structure of SACL-XLMR consists of a multilingual BERT (i.e., an embedding layer and  Transformer encoder) and a sentiment classifier.

\paragraph{Multilingual BERT} \label{sec:xlmr}
We apply a multilingual BERT model \citep{DBLP:conf/nips/ConneauL19,DBLP:conf/coling/AlabiAMK22}
on monolingual corpus to facilitate language adaptation. Besides, sentiment lexicon knowledge for each language is used to enhance sentiment-aware representation learning.

Formally, given an input token sequence
$x_{i1}, ..., x_{iN}$ where $x_{ij}$ refers to $j$-th token in the $i$-th input sample, and $N$ is the maximum sequence length, the model learns to generate the context representation of the input token sequences: 
\begin{equation}
\resizebox{0.89\linewidth}{!}{$
    \mathbf{h}_i = \text{BERT}(\texttt{[CLS]}, s_L, \texttt{[SEP]}, x_{i1}, ..., x_{iN}, \texttt{[SEP]}),
$}
\end{equation}
where \texttt{[CLS]} and \texttt{[SEP]} are special tokens, usually at the beginning and end of each sequence, respectively. $s_L$ refers to a token sequence of sentiment lexicon prefix corresponding to the input sequence.
$\mathbf{h}_i$ indicates the hidden representation of the $i$-th input sample, computed by the representation of \texttt{[CLS]} token in the last layer of the encoder.

\paragraph{Sentiment Classifier} \label{sec:senti}
Finally, according to the obtained representations, a sentiment classifier is applied to predict the sentiment label of each sample. 
\begin{equation}
    \hat{\mathbf{y}}_i = softmax(\mathbf{W}_h \mathbf{h}_i + \mathbf{b}_h),    
\end{equation} 
where $\mathbf{W}_h \in \mathbb{R}^{d_h \times |\mathcal{Y}|}$ and $\mathbf{b}_h \in \mathbb{R}^{|\mathcal{Y}|}$ 
are trainable parameters.
$|\mathcal{Y}|$ is the number of sentiment labels.

\subsection{Optimization Objective}
Supervised contrastive learning (SCL) \citep{DBLP:conf/nips/KhoslaTWSTIMLK20,DBLP:conf/iclr/GunelDCS21} is utilized to learn a generalized feature representation by capturing similarities between examples within a class and contrasting them with examples from other classes. However, directly compressing the feature space of each class can harm fine-grained features, which limits the model's ability to generalize. 
Recently, a new technique named supervised adversarial contrastive learning (SACL) \citep{hu2023supervised} has been proposed to address this issue by learning class-spread structured representations. The SACL uses both original and adversarial samples to effectively utilize prior information on label consistency and retain fine-grained features.

In this task, we apply the SACL technique to learn sentiment-spread representations and enhance the generalization of multilingual BERT. 
Formally, let us denote $I$ as the set of samples in a batch.
Define $\phi(i) = \{ e \in I \backslash \{i\}: \hat{\mathbf{y}}_e =\hat{\mathbf{y}}_i \}$ as the set of indices of all positives in the batch distinct from $i$, and $|\phi(i)|$ is its cardinality. 
The loss function of soft SCL is a weighted average of CE loss and SCL loss with a trade-off scalar parameter $\lambda$, i.e., 
\begin{equation}
\mathcal{L}_{\text{soft-SCL}} = \mathcal{L}_{\text{CE}} + \lambda \mathcal{L}_{\text{SCL}},  \label{eq:softscl}
\end{equation}
where 
\begin{equation}
\mathcal{L}_{\text{CE}} = -   \sum\limits_{i \in I} {\mathbf{y}}_{i,k} \log (\hat{\mathbf{y}}_{i,k}), 
  \end{equation}   
\begin{equation}
\resizebox{0.99\linewidth}{!}{$
    \mathcal{L}_{\text{SCL}} = \sum\limits_{i \in I} \frac{-1}{|\phi(i)|}  \sum\limits_{e \in \phi(i)} \log \frac{\exp(sim(\mathbf{z}_i,\mathbf{z}_e) / \tau)  } {\sum\limits_{a\in A(i)} \exp(sim(\mathbf{z}_i, \mathbf{z}_a) / \tau) }.
    $}
\end{equation}
$\mathbf{y}_{i,k}$ and $\hat{\mathbf{y}}_{i,k}$ denote the value of one-hot vector $\mathbf{y}_{i}$ and probability vector $\hat{\mathbf{y}}_{i}$ at class index k, respectively.
$A(i) = I \backslash \{i\} $. 
$\mathbf{z}_i = \mathbf{W}_h \mathbf{h}_i + \mathbf{b}_h$.
$sim(\cdot, \cdot)$ is a pairwise similarity function, 
i.e., dot product.
$\tau > 0$ is a scalar temperature parameter that controls the separation of classes. 

\begin{table*}[t]
\centering
\resizebox{\linewidth}{!}{
\begin{tabular}{l|r|r|r|c|c|l} 
\hline
\multicolumn{1}{c|}{\multirow{1}{*}{Model}} 
& \multicolumn{1}{c|}{\multirow{1}{*}{\# Param.}} 
& \multicolumn{1}{c|}{\multirow{1}{*}{\# Vocab.}}
& \multicolumn{1}{c|}{\multirow{1}{*}{\# Lang.}} 
& \multicolumn{1}{c|}{\multirow{1}{*}{Seen Lang. Adapt.}}
& \multicolumn{1}{c|}{\multirow{1}{*}{Unseen Lang. Adapt.}} 
& \multicolumn{1}{c}{\multirow{1}{*}{Lang. supported in AfriSenti datasets}} \\ 
\hline
XLM-R               & 270M  & 250k & 100 &  \xmark & \xmark 
& \texttt{amh}, \texttt{arq}, \texttt{hau}, \texttt{ary}, \texttt{pt-MZ}, \texttt{orm}, \texttt{swa} \\ 
\hline
AfriBERTa           & 126M  &  70k  & 11  & \cmark  & \xmark
& \texttt{amh}, \texttt{hau}, \texttt{ibo}, \texttt{kin}, \texttt{pcm}, \texttt{orm}, \texttt{swa},  \texttt{tir}, \texttt{yor} 
\\ \hline
AfroXLMR & 270M  & 250k  & 20  & \cmark & \xmark
& \texttt{amh}, \texttt{arq}, \texttt{hau}, \texttt{ibo}, \texttt{kin}, \texttt{ary}, \texttt{pcm}, \texttt{orm}, \texttt{swa}, \texttt{yor} \\ 
\hline
\textbf{SACL-XLMR} & 270M  & 250k  & 20  & \cmark  & \cmark
& \texttt{amh}, \texttt{arq}, \texttt{hau}, \texttt{ibo}, \texttt{kin}, \texttt{ary}, \texttt{pcm}, \texttt{orm}, \texttt{swa}, \texttt{yor} \\ 
\hline
\end{tabular}
}
\caption{
Comparison of our SACL-XLMR with other PTMs. \# Param. refers to the total number of parameters for each model excluding the task-specific classifier. 
\# Vocab. represents the size of vocabulary. 
\# Lang. indicates the number of language coverage. 
Seen/Unseen Lang. Adapt. represents whether the model supports seen/unseen target language adaptation. 
We list the languages covered by both the pre-trained corpus and AfriSent datasets.
}
\label{tab:language}
\end{table*}

At each step of training, under the soft SCL objective, we apply an adversarial training strategy (e.g., FGM \citep{DBLP:conf/iclr/MiyatoDG17}) on original samples to generate adversarial samples.
These samples can be seen as hard positive examples, which spread out the representation space for each sentiment class and confuse robust-less models.
After that, we utilize a new soft SCL on obtained adversarial samples to maximize the consistency of sentiment-spread representations with the same sentiment label.
Following the above calculation process 
of $\mathcal{L}_{\text{soft-SCL}}$ 
on original samples, the optimization objective on corresponding adversarial samples can be easily obtained in a similar way, i.e., $\mathcal{L}_{\text{soft-SCL}}^{\text{r-adv}}$.

The overall loss of SACL is defined as a sum of two soft SCL losses on both original and adversarial samples, i.e.,
\begin{equation}
     \mathcal{L} = \mathcal{L}_{\text{soft-SCL}} + \mathcal{L}_{\text{soft-SCL}}^{\text{r-adv}}. 
\end{equation}

\section{Experimental Setup}
\subsection{Comparison Methods}
We compare SACL-XLMR with the following several methods:
\begin{itemize}
    \item \textbf{Random} is based on random guessing, choosing each class/label with an equal probability.
    \item \textbf{XLM-R} \citep{DBLP:conf/nips/ConneauL19} is a multilingual variant of RoBERTa \citep{DBLP:journals/corr/abs-1907-11692}. It is pre-trained on filtered CommonCrawl data containing 100 languages.
    We use \textit{xlm-roberta-base}\footnote{\url{https://huggingface.co/}\label{code}} to initialize XLM-R. 
    \item \textbf{AfriBERTa} \citep{ogueji-etal-2021-small} is an Afro-centri multilingual language model pretrained on 11 African languages. 
    It is trained on an aggregation of datasets from the BBC news website and Common Crawl.
    We use \textit{castorini/afriberta\_large}\textsuperscript{\ref{code}} to initialize AfriBERTa. 
    \item \textbf{AfroXLMR} \citep{DBLP:conf/coling/AlabiAMK22} is an XLM-R model adapted to African languages. It is obtained by MLM adaptation of XLM-R on 17 African languages covering the major African language families and 3 high resource languages (Arabic, French, and English). 
    We use \textit{Davlan/afro-xlmr-large}\textsuperscript{\ref{code}} to initialize AfroXLMR. 
\end{itemize}

We report the comparison of our SACL-XLMR and the above PTMs in Table~\ref{tab:language}.

\subsection{Implementation Details}

\begin{table}[t]
\centering
\begin{tabular}{l|c}
\hline 
\multicolumn{1}{c|}{Hyperparameter}    &  SACL-XLMR  \\ \hline
Hidden size \textbf{$d_u$} & $1024$ \\ 
Perturbation radius   &  $\{0.5, 5\}$ \\  
Perturbation rate  & $\{0.1, 1\}$  \\
Trade-off weight $\lambda$ and $\lambda^{\text{r-adv}}$  & $\{0.05, 0.1\}$ \\ 
Temperature $\tau$ and $\tau^{\text{r-adv}}$  & $0.1$ \\ %
\hline 
Number of epochs & $10$ \\
Patience & $3$ \\ 
Batch size & $128$ \\
Learning rate   & $1e^{-5}$  \\ 
Weight decay & $1e^{-2}$  \\ 
Dropout & $0.2$ \\
Maximum token length & $250$ \\
\hline
\end{tabular}
\caption{Hyperparameter settings of SACL-XLMR.}
\label{tab:param}
\end{table}

All experiments are conducted on a single NVIDIA Tesla V100 32GB card. 
Stratified k-fold cross validation  \citep{DBLP:conf/ijcai/Kohavi95} is performed to split combined training and validation data of 12 African languages into 5 folds. Train/validation sets for Oromo (\texttt{orm}) and Tigrinya (\texttt{tir}) are not used due to the limited size of the data. We only evaluate on them in a zero-shot transfer setting.
We choose the optimal hyperparameter values based on the the average result of validation sets for all folds, and evaluate the performance of our system on the test data.
Following the scoring program of AfriSenti-SemEval, we report the weighted-F1 (w-F1) score to measure the overall performance.

Our SACL-XLMR is initialized with the \textit{Davlan/afro-xlmr-large}\textsuperscript{\ref{code}} parameters, due to the nontrivial and consistent performance in both subtasks.
The network parameters are optimized by using Adam optimizer \citep{DBLP:journals/corr/KingmaB14}.
The class weights in CE loss are applied to alleviate the class imbalance problem 
and are set by their relative ratios in the train and validation sets.
The detailed experimental settings on both two subtasks are in Table~\ref{tab:param}.

To effectively utilize sentiment lexicons of partial languages in
the AfriSenti datasets, we concatenate the corresponding lexicon prefix with the original input text. Given the $i$-th input sample, the lexicon prefix can be represented as 
$y_k: w_{k1}, ..., w_{kM}$ where $y_k$ is the sentiment label, $w_{km}$ refers to the corresponding $m$-th lexicon token in the original sequence. 
For our final system, we only use sentiment lexicons on the zero-shot subtask. We do not use it on the multilingual subtask due to the fact that some languages in the multilingual target corpus do not have available sentiment lexicons, making it difficult for the model to adapt effectively.

\section{Results and Analysis}

\subsection{Overall Results}
The overall results  for both subtasks are summarized in Table~\ref{tab:result_1} and \ref{tab:result_2}. 
From the results, it is not surprising that all pre-trained models clearly outperformed the Random baseline. The proposed SACL-XLMR consistently outperformed the comparison methods on both subtasks. 
Specifically, SACL-XLMR achieved \textbf{1.1\%} and \textbf{2.8\%} absolute improvements on the multilingual and zero-shot sentiment classification subtasks, respectively.

Moreover, we present the official results from
several top-ranked systems for the zero-shot sentiment classification subtask in AfriSenti-SemEval Shared Task (i.e., SemEval-2023 Task 12) in Table~\ref{tab:result_3}.
Our submitted system obtained the \textbf{1st} overall rank on the zero-shot sentiment classification subtask in the official ranking.

\begin{table}[t]
\centering
\begin{tabular}{l|c}
\hline
\multicolumn{1}{c|}{\multirow{1}{*}{Model}} & \texttt{multilingual} \\
\hline
Random      & 33.3 \\ 
XLM-R       & 62.5 \\
AfriBERTa   & 64.5 \\
AfroXLMR    & 69.9 \\
\hline
SACL-XLMR$_{fold1}$$^\dagger$     & 70.3 \\ 
\textbf{SACL-XLMR}  & \textbf{71.0} \\ 
Improve     & +1.1\% \\
\hline
\end{tabular}
\caption{
Experimental results (\%) against various methods on the multilingual sentiment classification subtask.
We present the weighted-F1 score to measure the performance.
All compared pre-trained models are fine-tuned on the multilingual dataset.
${fold1}$ means the result using only training data of one fold.
$^\dagger$ indicates the results on the official ranking.
}
\label{tab:result_1}
\end{table}

\begin{table}[t]
\centering
\resizebox{\linewidth}{!}{
\begin{tabular}{l|cc|c}
\hline
\multicolumn{1}{c|}{\multirow{1}{*}{Model}} 
 & \texttt{tir} & \texttt{orm}  & \multicolumn{1}{c}{\multirow{1}{*}{Avg.}}  \\ 
\hline
Random      & 34.3 & 33.6 & 34.0 \\ 
XLM-R       & 43.8 & 35.9 & 39.9 \\
AfriBERTa   & 44.1 & 43.6 & 43.9 \\
AfroXLMR    & 69.8 & 42.3 & 56.1 \\
\hline
SACL-XLMR$_{fold1}$$^\dagger$    & 70.5 & 45.8 & 58.2 \\ 
\textbf{SACL-XLMR}  & \textbf{71.8} & \textbf{46.0} & \textbf{58.9} \\
Improve     & +2.0\%    & +2.4\%    & +2.8\%\\
\hline
\end{tabular}
} 
\caption{
Experimental results (\%) against various methods on the zero-shot sentiment classification subtask.
We present the weighted-F1 score to measure the performance.
All compared pre-trained models are fine-tuned on the multilingual dataset.
${fold1}$ means the result using only training data of one fold.
$^\dagger$ indicates the results on the official ranking.
} 
\label{tab:result_2}
\end{table}

\begin{table}[t]
\centering
\resizebox{\linewidth}{!}{
\begin{tabular}{c|l|cc|c}
\hline
\multicolumn{1}{c|}{\multirow{1}{*}{Overall Rank}} & \multicolumn{1}{c|}{\multirow{1}{*}{Team Name}} 
 & \texttt{tir} & \texttt{orm}  & \multicolumn{1}{c}{\multirow{1}{*}{Avg.}}  \\ 
\hline
\textbf{Top 1} & \textbf{UCAS-IIE-NLP} & 70.47	& 45.82	& \textbf{58.15} \\ 
Top 2 & BCAI-AIR3  & \textbf{70.86} & 44.97 & 57.92 \\
Top 3 & ymf924	& 70.39	& 45.34	& 57.87 \\
- & UM6P    & 69.53	& 45.27	& 57.40 \\
- & TBS     & 69.61	& 45.12	& 57.37 \\
- & uid     & 69.90	& 44.75	& 57.33 \\
- & mitchelldehaven & 66.96	& \textbf{46.23}	& 56.60 \\
\hline
\end{tabular}
}
\caption{
Results of our submitted system compared with several top-ranked systems for the zero-shot sentiment classification subtask in AfriSenti-SemEval Shared Task.
The official scoring program uses the weighted-F1 score to measure the performance.
Following the AfriSenti SemEval Prizes\protect\footnotemark[1] and the task description \citep{muhammadSemEval2023}, the overall rank is calculated by averaging the results of all the languages in the subtask.
} 
\label{tab:result_3}
\end{table}

\subsection{Ablation Study}

\begin{figure*}[t]
    \centering
    \includegraphics[width=0.78\linewidth]{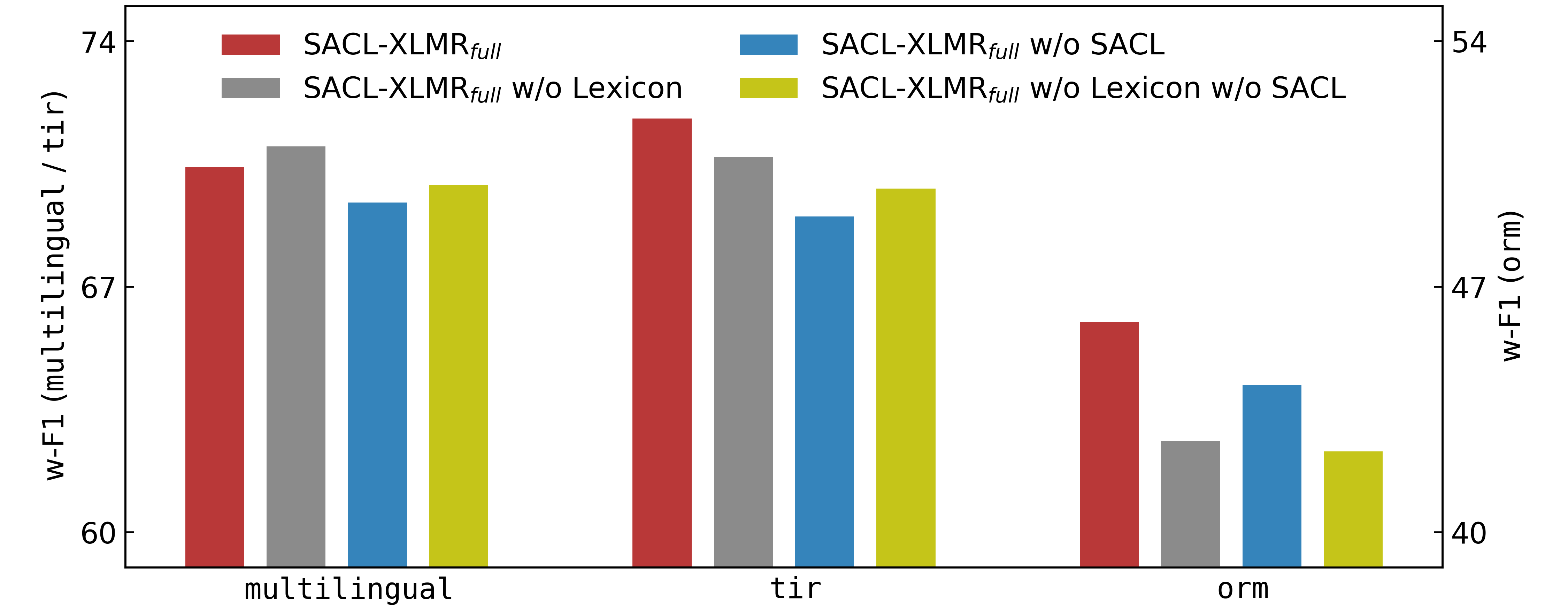}
    \caption{Ablation study results on two subtasks. We report the weighted-F1 score.     }
    \label{fig:result_abla}
\end{figure*}

\begin{figure*}[t]
\centering
\includegraphics[width=0.99\linewidth]{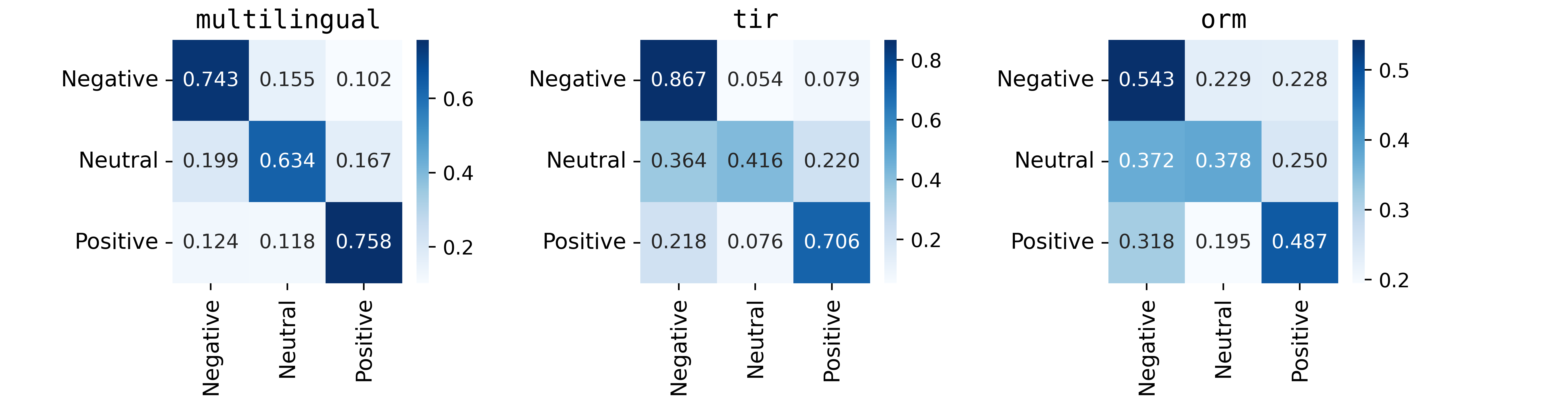}
\caption{The normalized confusion matrices for SACL-XLMR on three test sets of AfriSenti.
The rows represent the actual sentiment labels, whereas the columns represent predictions made by the model. 
Each cell $(i,j)$ represents that the percentage of class $i$ was predicted as class $j$.
The values of the diagonal elements represent the degree of correctly predicted classes. 
The higher the diagonal values of the confusion matrix the better, indicating many correct predictions.    
}
    \label{fig:error}
\end{figure*}

In this part, we conduct ablation studies by removing key components of SACL-XLMR$_{full}$ to further understand the proposed model:
\begin{itemize}
    \item  \textbf{- w/o Lexicon} refers to removing the sentiment lexicon.
    \item  \textbf{- w/o SACL} is an ablated model removing the supervised adversarial contrastive learning objective.
    \item  \textbf{- w/o Lexicon - w/o SACL} indicates removing both sentiment lexicons and SACL objective, degenerated to AfroXLMR.
\end{itemize}

Figure~\ref{fig:result_abla} shows results of ablation studies on two subtasks for low-resource sentiment analysis. 
Our SACL-XLMR$_{\text{w/o Lexicon}}$ and
SACL-XLMR$_{full}$  yield
the best performance on multilingual and zero-shot sentiment classification subtasks, respectively. When removing SACL objective, the results consistently decline on all subtasks, showing the effectiveness of SACL.

For the multilingual sentiment classification subtask, SACL-XLMR$_{full}$ obtains sub-optimal  results. This is most likely due to the fact that some languages in the target corpus do not have available sentiment lexicons, making it difficult for the model to adapt effectively.
Also, another caused factor is the incompleteness and poor quality of lexicon.
For the zero-shot sentiment classification subtask, the SACL-XLMR$_{full}$ yields the best performance on both \texttt{tir} and \texttt{orm} languages. 
It shows the effectiveness of sentiment lexicons in zero-shot scenarios, even if its quality is not good enough.

\subsection{Error Analysis}

Figure~\ref{fig:error} shows an error analysis of our system on two subtasks of AfriSenti-SemEval, including a multilingual test set and two zero-shot test sets. The normalized confusion matrices are used to evaluate the quality of the predicted outputs of SACL-XLMR.

From the diagonal elements of the matrices, 
true positives of non-neutral labels exceed those of the neutral label.
The results show that positive and negative features are more likely to adapt to low-resource languages. 
Besides, the above phenomenon is more obvious for \texttt{tir} and \texttt{orm} languages.
It indicates that SACL-XLMR can further facilitate language adaptation for low-resource languages by making full use of existing sentiment lexicons which contain only positive and negative words.

The confusion matrix of SACL-XLMR reveals the most confusing pair of sentiment labels: neutral to negative, especially for \texttt{tir} and \texttt{orm} languages in a zero-shot setting. 
The performance on \texttt{orm} language is relatively poor. Apart from the complexity of the language and the phenomenon of data scarcity, it is also due to the significant differences between \texttt{orm} and other African languages.
Considering the above issues make the task optimization more difficult, there is still a lot of room for improvement.

\section{Conclusion}
In this paper, a multilingual system named SACL-XLMR has been proposed for sentiment analysis on low-resource African languages. The system employs a lexicon-based multilingual BERT to facilitate language adaptation and sentiment-aware representation learning. It also uses a supervised adversarial contrastive learning technique to learn sentiment-spread structured representations and enhance model generalization. The system achieved competitive results, largely outperforming the comparison baselines on both multilingual and zero-shot sentiment classification subtasks, and obtained the 1st rank on zero-shot classification subtask in the official ranking.

\section*{Acknowledgements}
All the work in this paper are conducted during the SemEval-2023 Competition.
We thank the SemEval-2023 organizers and AfriSenti-SemEval task organizers for making this research possible.
We also appreciate the anonymous reviewers for their insightful and constructive comments that have helped us improve the quality of the paper.

\bibliography{semeval23}
\bibliographystyle{acl_natbib}

\end{document}